\pdfoutput=1
\documentclass[runningheads]{llncs}
 
\usepackage[T1]{fontenc}
\usepackage{graphicx}
\usepackage{amsmath}
\usepackage{amssymb}
\usepackage{booktabs}
\usepackage{url}
\usepackage[hidelinks]{hyperref}
\usepackage{xcolor}
\usepackage{caption}

\usepackage{tikz}
\usetikzlibrary{arrows.meta, positioning, shapes.geometric, fit, calc}
\usepackage{pgfplots}
\pgfplotsset{compat=1.18}
 
\begin{document}
 
\title{Tokenized Reward Systems for Federated Learning: The Privacy vs. Fairness Tradeoff}
 
\titlerunning{Privacy--Fairness Trade-offs in FL Reward Systems}
 
\author{Andreas Sendros\inst{1,2}\orcidID{0000-0002-6337-4320} \and
Nikolaos Pavlidis\inst{1,2}\orcidID{0000-0001-9370-5023} \and
Vasilios Perifanis\inst{1,2}\orcidID{0000-0003-3915-9628} \and
Pavlos S.\ Efraimidis\inst{1,2}\orcidID{0000-0003-3749-0165}}
 
\authorrunning{A. Sendros et al.}
 
\institute{Department of Electrical and Computer Engineering, 
Democritus University of Thrace, 67100 Xanthi, Greece \\
\email{\{asendros,npavlidi,vperifan,pefraimi\}@ee.duth.gr}
\and
Institute for Language and Speech Processing, Athena Research Center, 67100 Xanthi, Greece \\
\email{\{asendros,nikolaos.pavlidis,vperifan,pefraimi\}@athenarc.gr}}
 
\maketitle
 
\begin{abstract}
Federated learning (FL) systems increasingly rely on blockchain based reward mechanisms to incentivize participation while preserving data privacy. We design and empirically evaluate two complementary Solidity smart contracts for FL reward distribution: the first one that allocates rewards proportionally to a publicly disclosed quality score, and a privacy preserving variant that hides individual quality scores behind a zero-knowledge (ZK) proof of a threshold condition. Both contracts mint a non transferable, use-and-burn token maintaining a trustless incentive structure. We evaluate the two designs through MNIST training with adversarial behavior, including Sybil attacks, zero gradient and random noise, across three independent random seeds.
 
Our results reveal a fundamental tradeoff between cryptographic privacy and incentive fairness. The privacy preserving architecture strictly degrades system robustness, doubling the adversary's average reward capture while introducing severe outcome variance. This vulnerability is driven by a structural self-inclusion bias within cosine-similarity scoring: as the global model converges, legitimate gradient updates naturally diminish, rendering them statistically indistinguishable from bounded noise. Adversaries exploit this dynamic to systematically breach the ZK threshold, triggering a late stage failure cascade that culminates in absolute adversary monopoly over the reward pool and the total exclusion of honest contributors. Alongside a closed-form theorem formalizing this privacy--Sybil resistance tradeoff, we map strict mainnet deployment boundaries via linear gas scaling models, explicitly incorporating the post-Fusaka EIP-7825 per transaction gas cap. Ultimately, while cryptographically secure, static threshold ZK gates fail as incentive layer fairness mechanisms under realistic FL convergence dynamics on standard benchmarks.
 
\end{abstract}
 
\section{Introduction}
\label{sec:intro}
 
Federated Learning (FL)~\cite{mcmahan_2017} has emerged as a powerful paradigm for training machine learning models collaboratively while preserving data privacy in various domains such as healthcare~\cite{li_2025}, telecommunications~\cite{pavlidis_2025} and recommendation systems~\cite{perifanis_2022}. Unlike traditional centralized learning, FL allows multiple clients to train local models on their private data, sharing only model updates with a central aggregation server. This decentralized nature of FL reduces privacy risks but introduces several challenges, including client selection~\cite{pavlidis_2023}, fairness in participation, and resource heterogeneity~\cite{kairouz_2021}.
One critical challenge that arises is ensuring sustained user participation. Since model training in FL requires computational resources, energy, and time, users need strong incentives to remain engaged in the federated learning process. Designing an effective incentive mechanism is critical to maintaining a stable and high-quality learning ecosystem~\cite{zeng_2021}.
 
Traditional reward systems in FL often rely on centralized payment schemes or simple contribution based incentives, but these approaches introduce privacy risks, unfair compensation, and speculation-prone reward models~\cite{zhan_2021}. In privacy critical applications, rewards must be designed to ensure security, fairness, and anonymity, protecting user contributions while maintaining system transparency. blockchain based incentive mechanisms have emerged as a potential solution, offering decentralized, trustless, and privacy preserving rewards. However, existing blockchain implementations face several challenges, including high transaction costs, limited scalability, and vulnerability to speculative behavior in tokenized rewards~\cite{zhu_2023}.
 
A natural evolution is leveraging zero-knowledge (ZK) proofs: clients submit a proof that their quality score exceeds a threshold $\tau$ without revealing the exact score, and a smart contract distributes rewards among verified clients~\cite{heiss_2022,ebrahimi_2024,zkhybridfl_2025}. While this design provides cryptographic privacy, its empirical robustness under actual FL convergence dynamics remains largely unexplored.
 
This work addresses this gap by designing and empirically evaluating a privacy preserving blockchain based incentive mechanism. We propose two Solidity smart contracts: \textsc{FLRewardSystem} (distributing rewards based on public quality scores) and \textsc{FLRewardSystemZK} (hiding scores via zk-SNARKs~\cite{ref_groth16}). Both utilize a non transferable, use-and-burn token to prevent speculation. We evaluate these designs through end-to-end MNIST training (LeNet-5, 30 clients, Dirichlet $\alpha = 0.5$, 20 rounds) under realistic adversarial behavior.
 
\paragraph{Contributions.} 
This paper's contributions can be summarized as follows:
\begin{itemize}
    \item \textbf{Two contracts and an end-to-end testbed.} We implement a standard quality weighted reward contract and a privacy preserving ZK variant in Solidity, both minting a non transferable token, and integrate them with a full FedAvg training loop on MNIST. This lets us observe gas costs and reward outcomes for both designs under identical, realistic conditions.
    \item \textbf{Gas profiling.} We measure gas consumption via Hardhat in both contracts, showing linear scaling. Factoring in the recent EIP-7825 per transaction gas cap, Single Transaction limits drop to 270 participants, mandating Layer-2 scaling for larger networks.
    \item \textbf{A bias in the cosine quality score.} We describe and empirically check a closed-form approximation (Eq.~\ref{eq:self-inclusion}) suggesting that the FedAvg cosine score tends to inflate noise adversary scores as the model converges.
    \item \textbf{A privacy--fairness tradeoff.} We analyze when the ZK adversary share can exceed the standard share (Section~\ref{sec:adversarial}) and find that, in our experiments, ZK adversary capture is about 2.4$\times$ higher than standard, with a gate failure pattern in which a persistent noise adversary can capture most of the late round rewards.
\end{itemize}
 
\paragraph{Paper organization.} 
Section~\ref{sec:background} reviews related work. Section~\ref{sec:design} details our contracts. Section~\ref{sec:gas} reports gas scalability. Section~\ref{sec:adversarial} mathematically models the privacy--Sybil trade-off, validated by the FL pipeline results in Section~\ref{sec:fl-pipeline}. Section~\ref{sec:discussion} discusses mitigations, and Section~\ref{sec:conclusion} concludes.
 
\section{Background and Related Work}
\label{sec:background}
 
Evaluating client contributions is a key challenge in FL. Several quality metrics have been proposed, such as Shapley values~\cite{liu_2024} and cosine similarity~\cite{shrestha_2025}. Although Shapley values provide a principled evaluation of marginal contributions, their high computational cost and volatility~\cite{geimer_2024} hinder their adoption in production environments. Consequently, distance-based metrics, primarily the cosine similarity between a client's gradient and the aggregate, are widely preferred as quality signals for both client selection~\cite{hu_2025} and reward weighting~\cite{shrestha_2025}.
 
Furthermore, to eliminate the need for centralized trust and automate reward distribution, recent frameworks integrate FL with smart contracts. For example, FedJudge~\cite{wang_2023} utilizes blockchain to transparently evaluate contributions. However, public on-chain evaluations introduce privacy risks by exposing client updates. To mitigate this, recent systems utilize cryptographic proofs. Heiss~et~al.~\cite{heiss_2022} and Federify~\cite{ebrahimi_2024} use zk-SNARKs for on-chain training verification, while ZK-HybridFL~\cite{zkhybridfl_2025} combines DAG ledgers with ZK-validated updates.
 
\textbf{Position of This Work.}
Examining the respective work that has been done, we observe that cosine-similarity quality scoring~\cite{shrestha_2025,hu_2025} and ZK-based reward gates~\cite{heiss_2022,ebrahimi_2024,zkhybridfl_2025} are widely used, but mostly studied in isolation. There are a few studies that analyze the gas costs of ZK-FL frameworks, but none of them research adversarial dynamics in late training stages~\cite{heiss_2022,ebrahimi_2024}. Similarly, other works focus purely on algorithmic Byzantine robustness (e.g., Krum~\cite{blanchard_2017}, Trimmed Mean~\cite{yin_2018}) without considering the associated on-chain reward economics. Current literature lacks an empirical evaluation of how standard quality metrics interact with privacy preserving threshold gates during actual FL training. This paper bridges that gap by examining the practical implications of combining these two technologies. Using theoretical incentive analysis, end-to-end FL simulations, and updated Ethereum gas profiling, we uncover a convergence-driven failure mode that emerges naturally from this integration.
 
\section{System Model and Proposed Approach}
\label{sec:design}
 
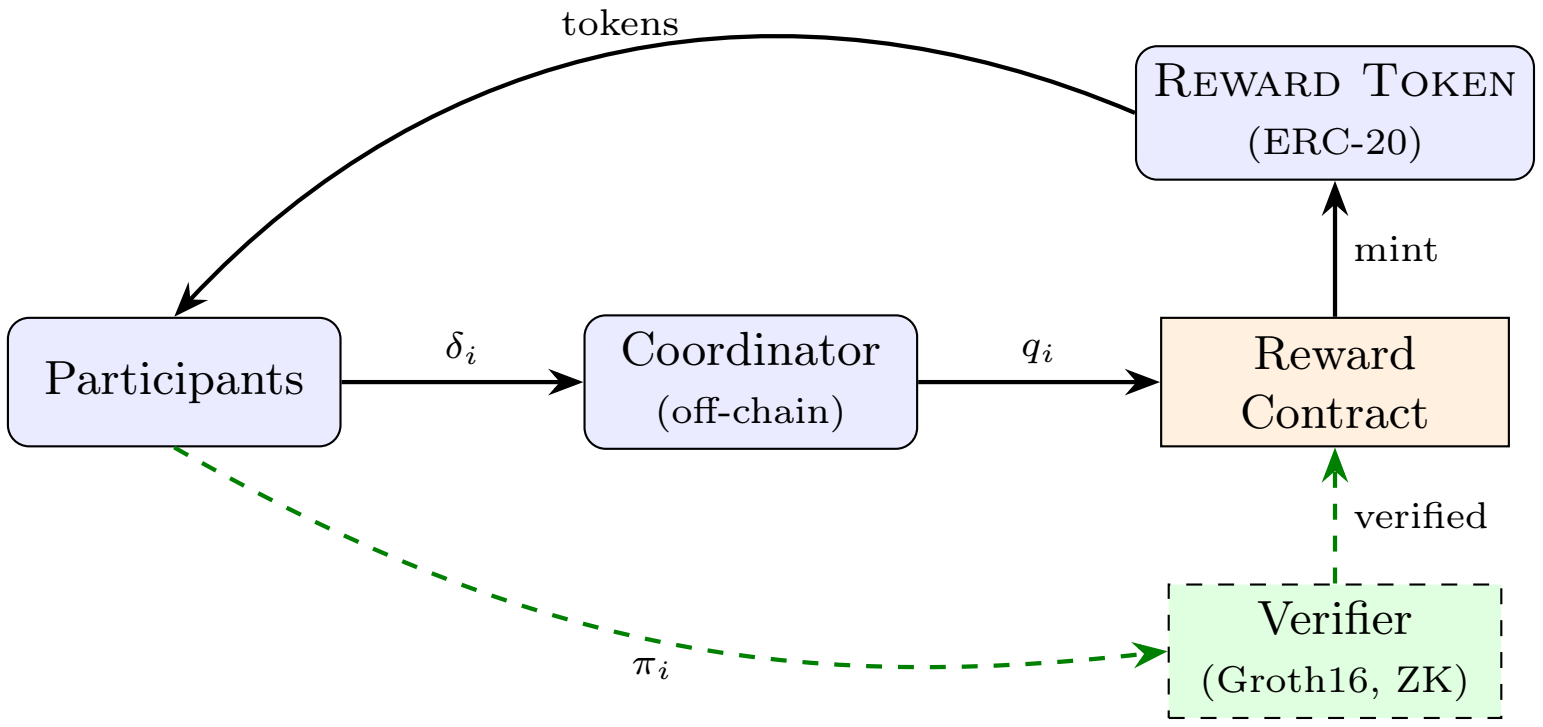
\begin{figure}[t]
\centering
\begin{tikzpicture}[
    node distance=0.9cm and 1.6cm,
    actor/.style={draw, rectangle, rounded corners, minimum width=2.2cm,
                  minimum height=0.85cm, align=center, font=\small,
                  fill=blue!8},
    contract/.style={draw, rectangle, minimum width=2.3cm,
                     minimum height=0.85cm, align=center, font=\small,
                     fill=orange!12},
    verifier/.style={draw, rectangle, minimum width=2.2cm,
                     minimum height=0.85cm, align=center, font=\small,
                     fill=green!12, dashed},
    arrow/.style={-Stealth, thick},
    arrowZK/.style={-Stealth, thick, dashed, draw=green!50!black}
]
 
\node[actor] (clients) {Participants};
\node[actor, right=of clients] (coord) {Coordinator \\ \scriptsize(off-chain)};
\node[contract, right=of coord] (reward) {Reward \\ Contract};
\node[actor, above=of reward] (flora) {\textsc{Reward Token} \\ \scriptsize(ERC-20)};
\node[verifier, below=of reward] (ver) {Verifier \\ \scriptsize(Groth16, ZK)};
 
\draw[arrow] (clients) -- node[above, font=\scriptsize] {$\delta_i$} (coord);
\draw[arrow] (coord) -- node[above, font=\scriptsize] {$q_i$} (reward);
\draw[arrow] (reward) -- node[right, font=\scriptsize] {mint} (flora);
\draw[arrow] (flora.west) to[bend right=35] 
    node[above, font=\scriptsize] {tokens} (clients.north);
 
\draw[arrowZK] (clients.south) to[bend right=18] 
    node[below, font=\scriptsize] {$\pi_i$} (ver.west);
\draw[arrowZK] (ver) -- node[right, font=\scriptsize] {verified} (reward);
 
\end{tikzpicture}
\caption{System architecture. The standard variant (solid arrows) distributes rewards via public scores $q_i$. The ZK variant (dashed arrows) hides $q_i$ using zero-knowledge proofs $\pi_i$ of $q_i \geq \tau$. Both mint non transferable reward tokens.}
\label{fig:reward-overview}
\end{figure}

To address the trust and privacy limitations of centralized contribution tracking, we propose a blockchain based reward architecture that issues non transferable tokens to verified participants. Figure~\ref{fig:reward-overview} illustrates the system.
 
\subsection{System Actors and Threat Model}
\label{sec:actors}
 
The architecture comprises four primary actor roles. The \textbf{coordinator} operates off-chain as a trusted entity that executes the Federated Averaging (FedAvg) algorithm, samples $K$ participants per round, calculates the cosine similarity quality scores $q_i$, and triggers the \texttt{distributeRewards} transaction. \textbf{Participants} are identified by on-chain ECDSA keys via the \texttt{registerParticipant} function. They train local models, submit gradient updates $\delta_i$ to the coordinator, and, in the ZK variant, generate cryptographic proofs of $q_i \geq \tau$. The \textbf{reward contract} (implemented in Solidity 0.8.20) manages the per-round reward pool, validates inputs, and mints the tokens. Finally, the \textbf{verifier contract} (exclusive to the ZK variant) acts as a pre-compiled Groth16 verifier, authenticating the $q_i \geq \tau$ proofs without exposing the actual $q_i$ values.
 
Regarding the threat model, we assume an honest coordinator alongside potentially adversarial participants. An adversary is capable of: (i)~executing Sybil attacks by registering up to $K$ malicious identities with equal stake; (ii)~acting as a free-rider by submitting zero gradients, random noise, or replaying stale updates; and (iii)~monitoring all public on-chain state. We assume adversaries cannot forge ECDSA signatures, break the underlying cryptographic primitives, or compromise the coordinator's infrastructure. Furthermore, we do not presume secure aggregation. The coordinator observes plaintext gradients, allowing our analysis to quantify residual incentive layer attacks under this baseline.
 
\subsection{Token Dynamics and Smart Contract Implementation}
\label{sec:token-dynamics}
 
The issued tokens operate as non transferable (``soulbound'') digital assets awarded exclusively for verified contributions. By utilizing an ERC-20 implementation where \texttt{transfer} and \texttt{transferFrom} unconditionally revert, the design neutralizes speculative behavior. Furthermore, tokens redeemed for ecosystem services are immediately burned via the \texttt{useTokens} function, ensuring they do not circulate beyond their intended utility. We note that mint and burn events remain linkable to participant addresses on-chain. The mechanism therefore provides speculation resistance rather than full transaction anonymity.
 
The allocation mechanism dynamically weights the expended computational resources ($S_i$) and the evaluated data quality ($Q_i \in [0, 100]$) relative to the per-round token budget ($\mathit{pool}$). The \texttt{distributeRewards} function mints rewards according to:
\begin{equation}
R_i = \frac{S_i \cdot Q_i \cdot \mathit{pool}}
{\left(\sum_j Q_j\right)\left(\sum_j S_j\right)}.
\label{eq:reward-std}
\end{equation}
Crucially, because the denominator aggregates across all registered participants, if only $K$ out of $N$ clients participate, only $\mathit{pool}/K$ of the total budget is minted. This intentional pool dilution property penalizes low participation rates.
 
\subsection{privacy preserving Variant: \textsc{FLRewardSystemZK}}
\label{sec:fl-reward-zk}
 
While the soulbound, use-and-burn model provides baseline privacy, the standard contract still publicly records individual quality scores $Q_i$ during reward distribution. To resolve this, \textsc{FLRewardSystemZK} integrates zero-knowledge proofs. Instead of revealing $q_i$, participants cryptographically prove that their score meets a configurable threshold ($q_i \geq \tau$). The contract only stores a binary \texttt{qualityVerified[address]} state. Rewards are subsequently distributed proportionally to the stake $S_i$ among the verified subset of clients $\mathcal{V}$:
\begin{equation}
R_i = \frac{S_i \cdot \mathit{pool}}{\sum_{j \in \mathcal{V}} S_j}, 
\qquad i \in \mathcal{V}.
\label{eq:reward-zk}
\end{equation}
 
Participants interact with the system by calling \texttt{submitQualityProof} from their Externally Owned Accounts (EOA). The contract triggers the Groth16 verifier and, upon successful verification, updates the \texttt{qualityVerified} mapping. Replay attacks are prevented via a \texttt{usedProofs} registry. The current circuit takes $q_i$ and randomness $r$ as private inputs and $commitment$ and $\tau$ as public inputs. We acknowledge that in this formulation the proof is not bound to a specific round identifier or a coordinator-authenticated digest of $\bar{\delta}^{(t)}$; a participant could in principle generate a valid proof for a self-chosen $q_i$ rather than the coordinator-assigned score. Binding the proof to a coordinator ECDSA signature over $(q_i, \texttt{round\_id}, \texttt{addr}_i)$ would close this gap at the cost of additional circuit complexity, and is left as future work. The \texttt{usedProofs} mapping prevents proof replay within a contract instance; including the round identifier in the commitment would extend this protection to cross-round replay. We compile the Groth16 circuit over the BN-254 curve using Circom~2.1, relying on the Powers of Tau ceremony for the trusted setup~\cite{ref_powersoftau,pinto_2020}. For the empirical evaluation (Section~\ref{sec:fl-pipeline}), we deploy a stub verifier preserving the exact contract interaction workflow while isolating verifier side pairing costs. This setup captures the exact contract side storage and execution overhead, excluding the pairing check gas costs, which are explicitly projected and analyzed in Section~\ref{sec:gas-end-to-end} and Section~\ref{sec:discussion}.
 
\section{Gas Cost and Scalability Analysis}
\label{sec:gas}
 
To evaluate deployment feasibility across Ethereum mainnet and Layer-2 scaling solutions, we benchmark the on-chain gas consumption of both reward contracts. Measurements were conducted using Hardhat~2.28.0 (Cancun hardfork, Solidity 0.8.20, optimizer enabled with 200 runs). All reported figures represent the median over 10 deterministic executions.
 
\subsection{Operation Level Baseline}
\label{sec:gas-operations}
 
Table~\ref{tab:op-baseline} details the baseline gas costs for core operations. Both contracts share identical costs for most functions, diverging primarily during \texttt{distributeRewards}. The ZK variant introduces a marginal overhead of 771~gas per participant (a 1.27\% increase over the standard contract's 60{,}738~gas). 
 
\begin{table}[t]
\caption{Per-operation gas consumption for \textsc{FLRewardSystem} 
(Standard) and \textsc{FLRewardSystemZK} (ZK), median of 10 runs.}
\label{tab:op-baseline}
\centering
\small
\begin{tabular}{lrrr}
\toprule
Operation & Standard & ZK & $\Delta$ \\
\midrule
\texttt{deploy}                 & 1{,}847{,}219 & 2{,}198{,}411 & +351{,}192 \\
\texttt{registerParticipant}    & 92{,}647 & 92{,}647 & 0 \\
\texttt{updateStake}            & 30{,}812 & 30{,}812 & 0 \\
\texttt{submitQualityProof}     & --- & 127{,}640 & --- \\
\texttt{distributeRewards}      & 60{,}738 & 61{,}509 & +771 \\
\quad (slope, gas / participant) & & & (+1.27\%) \\
\texttt{getReward}              & 28{,}453 & 28{,}453 & 0 \\
\texttt{transferStake}          & 53{,}218 & 53{,}218 & 0 \\
\bottomrule
\end{tabular}
\end{table}
 
This overhead comes from a cold SLOAD operation ($\sim$2{,}100~gas under EIP-2929) required to read the \texttt{qualityVerified} mapping. As demonstrated by recent comprehensive studies on EVM storage economics~\cite{kostamis_2024}, storage access operations constitute the primary bottleneck in smart contract scalability, a cost partially mitigated in our design by bypassing arithmetic calculations for unverified clients. Additionally, the ZK contract requires an extra 351~k gas during deployment, primarily due to the Groth16 verifier interface and the \texttt{usedProofs} registry.
 
\begin{figure}[t]
\centering
\begin{tikzpicture}
\begin{axis}[
    width=11cm, height=6cm,
    xlabel={Number of participants $N$},
    ylabel={\texttt{distributeRewards} gas (millions)},
    xmin=0, xmax=520,
    ymin=0, ymax=65,
    grid=major,
    grid style={dashed, gray!30},
    legend pos=north west,
    legend style={font=\small, draw=gray!50},
    label style={font=\small},
    tick label style={font=\footnotesize},
    every axis plot/.append style={very thick}
]
 
\addplot[blue, mark=*, mark size=2pt] coordinates {
    (50, 3.12) (100, 6.16) (250, 15.27) (500, 30.45)
};
\addlegendentry{Standard ($60{,}738 N + 82{,}253$)}
 
\addplot[red, mark=square*, mark size=2pt] coordinates {
    (50, 3.16) (100, 6.23) (250, 15.46) (500, 30.83)
};
\addlegendentry{ZK ($61{,}509 N + 80{,}300$)}
 
\addplot[dashed, gray, thick, no marks] coordinates {(0, 60) (520, 60)};
\addlegendentry{Post-Fusaka block limit (60M)}
 
\addplot[dotted, orange!80!black, thick, no marks] coordinates {(0, 16.78) (520, 16.78)};
\addlegendentry{EIP-7825 per-tx cap ($2^{24}$)}
 
\node[font=\footnotesize, anchor=south] at (axis cs:274, 17.3) {$N \approx 274$};
 
\end{axis}
\end{tikzpicture}
\caption{Linear gas scaling for \texttt{distributeRewards}. Dotted and dashed lines denote the EIP-7825 tx-cap (16.78M) and post-Fusaka block limit (60M), respectively. Single-tx capacity: 274 (Standard), 271 (ZK).}
\label{fig:gas scaling}
\end{figure}
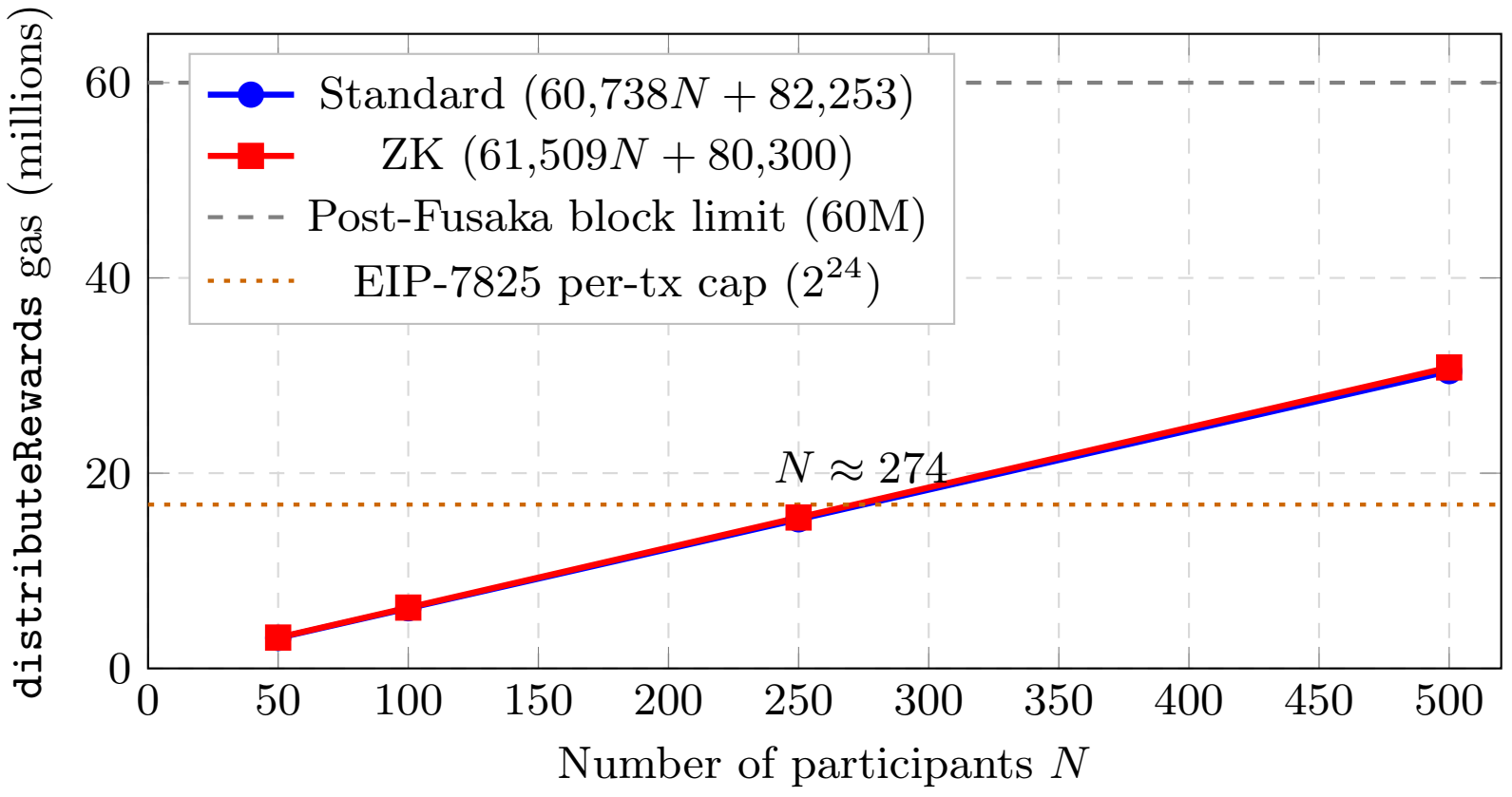

Scaling the network up to 5,000 participants reveals precisely linear gas consumption for the \texttt{distributeRewards} function (Fig.~\ref{fig:gas scaling}):
\begin{align}
\mathit{gas}_\text{Std} &= 82{,}253 + 60{,}738 \cdot N, \label{eq:gas-std}\\
\mathit{gas}_\text{ZK}  &= 80{,}300 + 61{,}509 \cdot N. \label{eq:gas-zk}
\end{align}
 
\subsection{Single Transaction Participant Limits}
\label{sec:gas-mainnet}
 
The December 2025 Fusaka hard fork fundamentally altered Ethereum's deployment constraints. While EIP-7935 doubled the block gas limit to 60~M, EIP-7825 simultaneously enforced a strict \emph{per transaction} cap of $2^{24} \approx 16.77$~M gas to mitigate DoS attacks and enable parallel execution~\cite{eip7825}. Consequently, EIP-7825 now acts as the absolute bottleneck for any single \texttt{distributeRewards} invocation. 
 
Applying this 16.77~M gas ceiling to Equations~\ref{eq:gas-std} and~\ref{eq:gas-zk} yields a maximum Single Transaction capacity of $N_\text{max,Std} = 274$ and $N_\text{max,ZK} = 271$ participants. Compared to the pre-Fusaka legacy limits (492 and 486, respectively), this represents a severe 44\% capacity reduction. Consequently, FL deployments exceeding $\sim$270 participants strictly require batched transactions or Layer-2 architectures. To our knowledge, this is the first analysis of FL incentive scalability to explicitly incorporate the EIP-7825 constraint.
 
\subsection{End-to-End Gas Cost: Standard vs.\ ZK}
\label{sec:gas-end-to-end}
 
Table~\ref{tab:gas-comparison} aggregates the cumulative gas overhead for our 20-round FL training pipeline. The Standard contract required 13.0~M gas total. The ZK variant, using the stub verifier, consumed 16.2~M gas, primarily due to the repeated \texttt{submitQualityProof} invocations (112~k--128~k gas per proof). Transitioning to a production ZK deployment with a full BN-254 Groth16 verifier introduces an estimated 175~k gas pairing check overhead per proof~\cite{ref_groth16}. This pushes the projected pipeline total to 29.8~M gas, representing a 2.29$\times$ overhead compared to the Standard model.
 
\begin{table}[t]
\caption{End-to-end gas consumption over the 20-round FL pipeline.}
\label{tab:gas-comparison}
\centering
\small
\begin{tabular}{lrrr}
\toprule
Component & Phase~2 Std & Phase~3 ZK (stub) & Phase~3 ZK (real) \\
\midrule
\texttt{distributeRewards} (total) & 13{,}028{,}590 & 7{,}210{,}890 & 7{,}210{,}890 \\
\texttt{submitQualityProof} (total) & 0 & 8{,}946{,}684 & 22{,}596{,}684 \\
\midrule
\textbf{Total gas (20 rounds)} & \textbf{13{,}028{,}590} & \textbf{16{,}157{,}574} & \textbf{29{,}807{,}574} \\
Multiplier vs.\ Standard & 1.0$\times$ & 1.24$\times$ & 2.29$\times$ \\
Per-round average & 651{,}430 & 850{,}399 & 1{,}568{,}820 \\
\bottomrule
\end{tabular}
\end{table}
 
\section{Adversarial Robustness}
\label{sec:adversarial}
 
Evaluating the robustness of both the Standard and ZK reward variants requires formalizing their behavior under Sybil and free-rider attacks. Let $K$ denote the number of Sybil identities controlled by a single adversary, and $n$ the number of honest participants. We define $Q_s$ and $Q_h$ as the quality scores assigned to a Sybil and an honest participant, respectively. To isolate the incentive dynamics, we assume equal computational stake across all clients ($S_i = 1$), aligning with standard FL benchmarks.
 
\subsection{Sybil Attacks: Closed-Form Reward Shares}
\label{sec:sybil-closed-form}
 
\paragraph{Standard system.}
Under the baseline system, rewards scale proportionally to the exposed quality score. Aggregating the $K$ Sybil identities against the $n$ honest participants, the adversary's captured fraction of the token pool is deterministic:
\begin{equation}
\mathit{adv\_share}_\text{Std} = 
\frac{K \cdot Q_s}{n \cdot \overline{Q_h} + K \cdot Q_s},
\label{eq:adv-share-std}
\end{equation}
where $\overline{Q_h}$ represents the mean quality of honest participants.
 
\paragraph{ZK system.}
Conversely, \textsc{FLRewardSystemZK} masks individual quality variations. Provided a client passes the zero-knowledge threshold proof ($Q_i \geq \tau$), the contract distributes rewards strictly by stake across the verified set $\mathcal{V}$. Assuming the adversary calibrates the Sybil nodes to just pass the threshold, their captured fraction becomes:
\begin{equation}
\mathit{adv\_share}_\text{ZK} = \frac{K}{n + K}.
\label{eq:adv-share-zk}
\end{equation}
By design, cryptographic privacy decouples the reward denominator from the actual quality contributions. This architectural choice introduces a fundamental tradeoff, which we empirically evaluate in Section~\ref{sec:fl-pipeline}.
 
Analytically comparing Equations~\ref{eq:adv-share-std} and~\ref{eq:adv-share-zk} reveals that the ZK variant allocates a larger share to the adversary strictly when the honest average quality exceeds the Sybil quality:
 
\begin{theorem}[Privacy--Sybil resistance trade-off]
\label{thm:privacy-sybil}
For equal-stake participants with $Q_s, \overline{Q_h} \geq \tau$,
\begin{equation*}
\mathit{adv\_share}_\textnormal{ZK} > \mathit{adv\_share}_\textnormal{Std} 
\quad \Longleftrightarrow \quad \overline{Q_h} > Q_s.
\end{equation*}
\end{theorem}
 
\begin{proof}
$\frac{K}{n+K} > \frac{KQ_s}{n\overline{Q_h} + KQ_s}$ iff 
$K(n\overline{Q_h} + KQ_s) > KQ_s(n+K)$ iff 
$Kn\overline{Q_h} + K^2 Q_s > KQ_s n + K^2 Q_s$ iff 
$Kn\overline{Q_h} > KQ_s n$ iff $\overline{Q_h} > Q_s$. \qed
\end{proof}
 
The condition $\overline{Q_h} > Q_s$ is satisfied in any functional FL environment where honest updates outperform marginal, threshold-passing Sybil submissions. Consequently, whenever the quality gate successfully discriminates high from low quality, enforcing ZK privacy strictly degrades Sybil resistance. Using a realistic baseline of $n = 20$ honest participants ($\overline{Q_h} \approx 75$), if a single Sybil adversary merely meets the threshold ($Q_s = 50$), the ZK system inflates their reward share by $1.4\times$ (4.8\% vs 3.3\% for Standard). With $K=10$ Sybils, the inflation is $1.3\times$ (33.3\% vs 25.4\%).
 
\subsection{Free-Riders}
\label{sec:free-riders}
 
Free-riders compromise system integrity by submitting zero effort gradients (e.g., pure noise, zeros, or stale replays). The standard system inherently penalizes this behavior: a zero-gradient yields $Q_s = 0$ (resulting in zero reward), while noise yields negligible similarity scores. 
 
In contrast, the ZK variant's binary gate imposes a severe incentive discontinuity. A free-rider scoring exactly at the threshold ($Q_s = \tau$) extracts the exact same $1/(n+1)$ reward share as an honest client providing an exceptional update ($Q_s \gg \tau$). As our empirical evaluation demonstrates (Section~\ref{sec:fl-pipeline}), random noise vectors systematically breach this static threshold during late stage training, allowing free-riders to capture rewards equivalent to honest participants under ZK, while remaining penalized under the Standard model.
 
\section{Empirical Validation: End-to-End FL Pipeline}
\label{sec:fl-pipeline}
 
To empirically evaluate the theoretical vulnerabilities identified in Section~\ref{sec:adversarial}, we deployed both \textsc{FLRewardSystem} and \textsc{FLRewardSystemZK} within an end-to-end federated training pipeline. The experimental setup comprises 30 clients training a LeNet-5 model ($\sim$44k parameters) on the MNIST dataset with non-IID partitions (Dirichlet $\alpha = 0.5$). The FedAvg aggregation runs for 20 rounds, sampling $K = 10$ clients per round performing 5 local SGD epochs (learning rate $0.01$, momentum $0.9$). We inject five adversaries into the pool: two zero-gradient free-riders, two random-noise free-riders ($\sigma = 0.01$), and one replay attacker submitting stale gradients from $t-5$. 
 
The coordinator calculates quality scores as $q_i = \lfloor 100 \cdot \max(0, \cos(\delta_i, \bar{\delta})) \rfloor$, with the ZK threshold strictly set to $q_i \geq 50$. All on-chain interactions are executed locally via Hardhat~2.28.0 (Cancun hardfork), and results are averaged across three independent random seeds (42, 43, 44).
 
\subsection{Cosine Self-Inclusion Bias}
\label{sec:self-inclusion}
 
\textbf{Finding 1.} The FedAvg cosine similarity metric suffers from a structural self-inclusion bias. When a client's update $\delta_i$ contributes to the aggregate $\bar{\delta}$ with weight $1/K$, its resulting score is artificially inflated:
\begin{equation}
\cos(\delta_i, \bar{\delta}) \approx 
\frac{\|\delta_i\|}{K \cdot \|\bar{\delta}\|}.
\label{eq:self-inclusion}
\end{equation}
 
Because the aggregate $\bar{\delta}$ inherently incorporates a $1/K$ fraction of $\delta_i$, the dot product between them is dominated by the adversary's self-term ($\frac{1}{K}\|\delta_i\|^2$). For high-dimensional random noise, the cross-terms between $\delta_i$ and the remaining updates vanish in expectation due to orthogonality. Normalizing this self-term by the respective vector magnitudes yields the approximation in Equation~\ref{eq:self-inclusion}.
 
We empirically verified this closed-form bias across all 11 rounds where a noise adversary was actively sampled (seed 42). As detailed in Table~\ref{tab:self-inclusion}, the measured $\cos(\delta_i, \bar{\delta})$ closely tracks the predicted baseline $\|\delta_i\| / (K \cdot \|\bar{\delta}\|)$ with a residual error strictly below 3\%. This confirms that self-inclusion, rather than actual alignment with the global objective, dictates the noise adversaries' scores.
 
\begin{table}[t]
\caption{Empirical confirmation of the cosine self-inclusion bias 
(Eq.~\ref{eq:self-inclusion}) for noise adversaries across 11 sampled 
rounds. Residual error $< 3\%$ in every measurement.}
\label{tab:self-inclusion}
\centering
\small
\begin{tabular}{rrrrrr}
\toprule
Round & $\|\delta_i\|$ & $\|\bar{\delta}\|$ & Predicted bias & Measured $\cos$ & Match \\
\midrule
2  & 2.10 & 1.73 & 0.121 & 0.117 & 96.7\% \\
4  & 2.11 & 0.84 & 0.252 & 0.243 & 96.6\% \\
5  & 2.12 & 0.68 & 0.310 & 0.303 & 97.6\% \\
7  & 2.11 & 0.65 & 0.323 & 0.328 & 98.5\% \\
10 & 2.10 & 0.43 & 0.491 & 0.496 & 98.9\% \\
12 & 2.11 & 0.42 & 0.506 & 0.513 & 98.6\% \\
15 & 2.10 & 0.35 & 0.594 & 0.596 & 99.7\% \\
17 & 2.11 & 0.40 & 0.524 & 0.527 & 99.4\% \\
18 & 2.11 & 0.31 & 0.682 & 0.681 & 99.9\% \\
19 & 2.13 & 0.36 & 0.589 & 0.593 & 99.4\% \\
20 & 2.11 & 0.33 & 0.649 & 0.640 & 98.6\% \\
\bottomrule
\end{tabular}
\end{table}
 
The practical impact of this bias is critical. As the global model converges, the aggregate norm shrinks ($\|\bar{\delta}\| \to 0$), whereas a noise adversary's norm remains statically bounded ($\|\delta_i\| \approx 2.1$, corresponding to the expected $L_2$ norm of $\mathcal{N}(0, 0.01^2)$ over $\sim$44k dimensions). Consequently, once $\|\bar{\delta}\|$ drops below $\approx 0.42$, the noise vector mathematically breaches the $\tau = 50$ ZK threshold. In our simulations, this threshold inversion occurs exactly at round 10. 
 
\begin{figure}[t]
\centering
\begin{tikzpicture}
\begin{axis}[
    width=11cm, height=6.5cm,
    xlabel={FL Round},
    ylabel={Mean quality score $q_i$},
    xmin=0, xmax=21,
    ymin=0, ymax=85,
    grid=major,
    grid style={dashed, gray!30},
    legend pos=north east,
    legend style={font=\small, draw=gray!50},
    label style={font=\small},
    tick label style={font=\footnotesize},
    every axis plot/.append style={very thick}
]
 
\addplot[blue, mark=*, mark size=2pt] coordinates {
    (1, 75.0) (2, 73.5) (3, 72.1) (4, 70.4) (5, 68.5)
    (6, 60.2) (7, 52.1) (8, 47.2) (9, 41.5) (10, 38.9)
    (11, 36.3) (12, 35.0) (13, 33.8) (14, 32.4) (15, 31.0)
    (16, 30.0) (17, 28.5) (18, 27.0) (19, 26.2) (20, 25.4)
};
\addlegendentry{Honest $\bar{q}_i$}
 
\addplot[red, mark=square*, mark size=2pt] coordinates {
    (1, 5.0) (2, 6.0) (3, 7.5) (4, 9.0) (5, 11.2)
    (6, 13.0) (7, 14.5) (8, 15.0) (9, 17.2) (10, 19.5)
    (11, 22.0) (12, 25.5) (13, 30.2) (14, 36.0) (15, 42.8)
    (16, 48.5) (17, 53.0) (18, 58.2) (19, 62.0) (20, 65.5)
};
\addlegendentry{Noise adversary $\bar{q}_i$}
 
\addplot[dashed, gray, thick, no marks] coordinates {(0, 50) (21, 50)};
\addlegendentry{ZK threshold $\tau = 50$}
 
\node[font=\footnotesize, anchor=west] at (axis cs:14.5, 60) {Crossover};
\draw[->, thin] (axis cs:14.8, 58) -- (axis cs:15.5, 47);
 
\end{axis}
\end{tikzpicture}
\caption{Temporal evolution of mean quality scores (seed 42). Honest $\bar{q}_i$ declines as the model converges, while noise adversary $\bar{q}_i$ rises due to self-inclusion bias (Eq.~\ref{eq:self-inclusion}). The ZK gate ($\tau=50$) preferentially admits adversaries after round 16.}
\label{fig:quality-decline}
\end{figure}
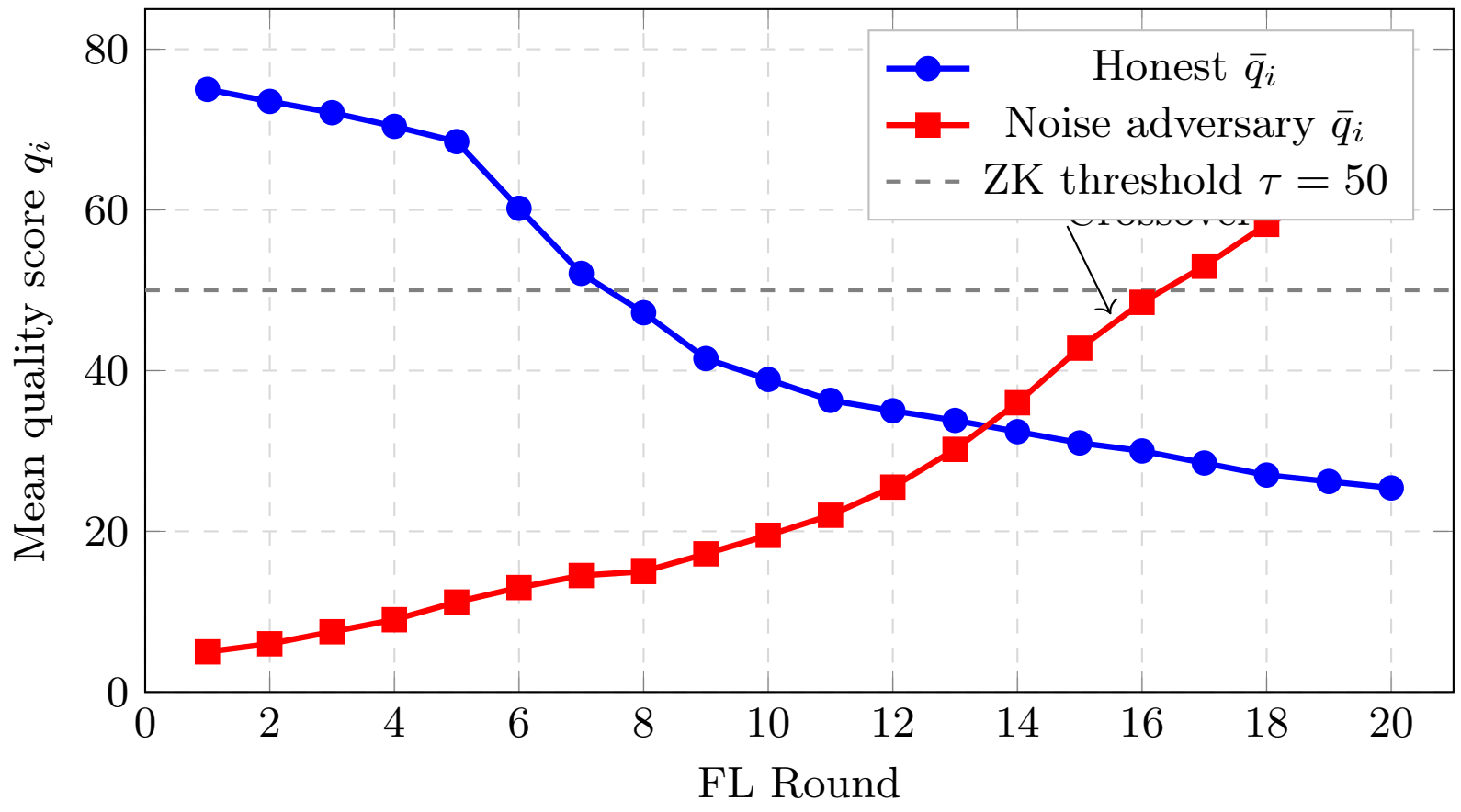

Legitimate clients suffer from the inverse dynamic: as their gradient updates naturally diminish upon convergence, their relative cosine scores collapse. Between rounds 1 and 5, the mean honest $q_i$ averages 72.1, but plummets to 28.0 by rounds 16--20. Simultaneously, the noise adversary's mean score surges from 6.1 to 48.8, ultimately surpassing the honest clients by round 16 (Figure~\ref{fig:quality-decline}).
 
\subsection{ZK Gate Failure Cascade}
\label{sec:gate-failure}
 
By bridging the FL rounds directly to the \textsc{FLRewardSystemZK} contract, we tracked the exact pass/fail rates of the $q_i \geq 50$ threshold gate. Table~\ref{tab:gate-cascade} details the resulting failure cascade observed in seed 42 (seeds 43 and 44 exhibit identical architectural trajectories).
During the final stages of training (rounds 17--20), the system enters a state of absolute adversary monopoly. The noise adversary ($q_i$ ranging from 53 to 68) becomes the exclusively verified participant, siphoning 100\% of the minted tokens across four consecutive rounds. The cryptographic gate, initially effective at filtering out malicious actors, actively enforces their monopoly during late convergence.
A secondary structural vulnerability emerges when no participants generate a valid proof, triggering an unconditional revert in \texttt{FLRewardSystemZK.distributeRewards()}. Across all tested seeds, this total gate failure occurred at least once (e.g., round 14 for seed 42, rounds 15--16 for seed 43). Managing this requires explicit pool rollovers, fallback mechanisms, or dynamic thresholding, representing a distinct operational limitation of static ZK gates.
 
\begin{table}[t]
\caption{ZK gate cascade in convergent FL (seed 42). $K = 10$ clients 
sampled per round, ZK threshold $q_i \geq 50$. Adversary monopoly 
emerges in rounds 17--20.}
\label{tab:gate-cascade}
\centering
\small
\begin{tabular}{lcccl}
\toprule
Round & $K_\text{verified}$ & Adv.\ verified & Honest verified & Outcome \\
\midrule
1--5    & 7--9 & 0 & 7--9 & Gate functions: only adversaries blocked \\
6--9    & 5--8 & 0 & 5--8 & Honest clients begin to fail \\
10--13  & 1--2 & 0--1 & 0--2 & Gate severely degraded \\
14      & \textbf{0}    & 0 & 0 & \textbf{Total gate failure: $K_\text{verified} = 0$} \\
15--16  & 1--2 & 0--1 & 1--2 & Partial recovery \\
\textbf{17--20}  & \textbf{1--2} & \textbf{1--2} & \textbf{0} & \textbf{Adversary monopoly: 100\% capture} \\
\bottomrule
\end{tabular}
\end{table}
 
\subsection{Multi-Seed Quantitative Comparison}
\label{sec:multi-seed}
 
Aggregating the outcomes across the three random seeds (Table~\ref{tab:multi-seed}) confirms the systemic degradation of fairness under the ZK architecture. The Standard contract successfully suppresses the adversary token capture below their proportional headcount (10.27\% $\pm$ 0.71\% actual vs.\ 16.67\% expected), demonstrating stable variance. 
 
In stark contrast, the ZK contract inflates the adversary's share to 24.51\% $\pm$ 6.93\%, effectively doubling their payout while introducing massive unpredictability (9.8$\times$ higher variance). Furthermore, the Gini coefficient for honest reward distribution deteriorates from 0.2171 to 0.4616. This proves that the static ZK gate not only fails to block adversaries in late rounds, but heavily skews the compensation among legitimate contributors.
 
\begin{table}[b]
\caption{Multi-seed outcomes (seeds 42, 43, 44). Standard limits 
adversary capture below headcount; ZK doubles adversary share, 
more than doubles honest reward inequality, and exhibits 
9.8$\times$ greater outcome variance.}
\label{tab:multi-seed}
\centering
\small
\begin{tabular}{lccc}
\toprule
Metric & Standard & ZK & Multiplier \\
\midrule
Adversary share (mean)    & 10.27\% & 24.51\% & 2.4$\times$ \\
Adversary share (std)     & $\pm$0.71\% & $\pm$6.93\% & 9.8$\times$ \\
Adversary share (range)   & 9.30--10.90\% & 15.09--31.58\% & --- \\
Honest reward Gini        & 0.2171 & 0.4616 & 2.1$\times$ \\
Skipped rounds (per 20)   & 0      & 1.33   & --- \\
Adversary monopoly rounds & 0      & 3.0    & --- \\
\bottomrule
\end{tabular}
\end{table}
 
\section{Discussion}
\label{sec:discussion}
 
Enforcing zero-knowledge thresholds over standard quality metrics fundamentally destabilizes reward distributions during late stage training, exposing a structural conflict between cryptographic privacy and incentive fairness.
 
\subsection{Why static threshold ZK Gates Fail}
\label{sec:why-fail}
 
This failure stems from the mechanics of FL convergence: honest gradient norms strictly decrease, while noise adversaries maintain a statistically constant gradient norm. Because cosine similarity normalizes by vector magnitudes, the shrinking aggregate norm artificially amplifies the relative weight of bounded orthogonal noise. The self-inclusion bias (Eq.~\ref{eq:self-inclusion}) therefore guarantees that any static threshold $\tau$ calibrated to admit early-stage honest updates will inevitably admit late stage noise. This architectural flaw applies to any \emph{magnitude relative} metric, consistently triggering massive outcome variance ($\pm 6.93\%$ adversary share) and ultimate adversary monopoly.
 
\subsection{Mitigations and Limitations}
\label{sec:mitigations}

Addressing this vulnerability requires decoupling quality evaluation from the decaying aggregate norm. \emph{Algorithmic adjustments} include leave one out aggregation ($\bar{\delta}_{-i}$), which eliminates self-inclusion bias but demands complex multi-pairing ZK circuits. Alternatively, Shapley-value rewards~\cite{liu_2022,liu_2024} evaluate marginal utility rather than magnitude, yet introduce severe computational overhead ($O(2^K)$) and require either public subset evaluations~\cite{ma_2021} or prohibitive ZK logic. Convergence-aware dynamic thresholds ($\tau^{(t)}$) preserve discriminative power but inflate proof complexity by requiring prior state commitments. 

More \emph{pragmatic defenses} include strict early stopping criteria, halting training before the aggregate norm shrinks sufficiently to admit bounded noise. Additionally, evaluating clients via historical quality tracking (e.g., exponential moving averages) heavily penalizes persistent noise during early rounds, neutralizing late stage threshold inversions. Beyond these algorithmic mitigations, richer defenses combining 
stake slashing penalties for detected malicious behavior, on-chain reputation tracking, or hybrid architectures integrating robust aggregation with ZK gates represent promising directions for future work.

Regarding \emph{limitations}, our pipeline utilized a stub ZK verifier to isolate contract side costs, though the $\sim$175~k gas Groth16 pairing check is factored into our final projections. Furthermore, while the mathematical bias (Eq.~\ref{eq:self-inclusion}) is universal, the exact crossover round is model-dependent, replicating these dynamics on heavier benchmarks (e.g., ResNet) will better map the failure cascade timeline. Finally, our threat model assumes an honest coordinator, leaving Byzantine score manipulation to broader frameworks like ZK-HybridFL~\cite{zkhybridfl_2025}. Additionally, while ZK proofs hide the exact $q_i$ value, metadata such as proof submission timing, reward mint events, and token burn patterns remain observable on-chain and could enable partial participation inference. Full participation-level privacy would require complementary techniques such as mixnets or timing  obfuscation, which we consider orthogonal to the incentive fairness question studied here.
 
\section{Conclusion}
\label{sec:conclusion}
 
This work evaluated the intersection of cryptographic privacy and incentive fairness in Federated Learning. Deploying \textsc{FLRewardSystem} and \textsc{FLRewardSystemZK} on MNIST under non-IID conditions, we demonstrated that static threshold ZK gates inherently fail under realistic convergence dynamics. 
 
We proved that FedAvg cosine-similarity scoring suffers from a structural self-inclusion bias ($\cos(\delta_i, \bar{\delta}) \approx \|\delta_i\| / (K \|\bar{\delta}\|)$) that systematically elevates bounded noise adversaries as the global model converges. Consequently, the ZK variant exhibited a 2.4$\times$ increase in adversary token capture and a 9.8$\times$ spike in outcome variance compared to the Standard baseline. The static ZK gate ultimately failed across all seeds, resulting in frequent contract reverts ($K_\text{verified} = 0$) and guaranteed adversary monopolies during late stage training.
 
Simultaneously, we mapped deployment limits via strict linear gas scaling models. Applying the post-Fusaka EIP-7825 per transaction gas cap restricts Single Transaction distributions to approximately 270 participants. Given this hard constraint alongside the projected 2.29$\times$ overhead for production Groth16 verification, Layer-2 scaling becomes mandatory for ZK-FL reward systems.
 
Ultimately, deploying static zero-knowledge thresholds over magnitude relative metrics fundamentally breaks incentive fairness. In our immediate future work, we plan to design and deploy zero-knowledge circuits that natively support leave one out aggregation and historical reputation tracking, aiming to definitively resolve the privacy--fairness conflict in tokenized FL environments.
 
\begin{credits}
\subsubsection{\ackname} 
This work has been partially supported by the EU project NGI Trustchain (grant No.\ 101093274) and partially by the project ``Digital Platform for Managing Scientific Conferences'' (code: PAMTHR05-0162478). The authors acknowledge the use of large language models for refining text and polishing grammar in this manuscript.
 
\subsubsection{\discintname}
The authors have no competing interests to declare that are relevant 
to the content of this article.
\end{credits}
 
\bibliographystyle{splncs04}

\end{document}